\documentclass[10pt,twocolumn,letterpaper]{article}
\usepackage{iccv}
\usepackage{times}
\usepackage{epsfig}
\usepackage{graphicx}
\usepackage{amsmath}
\usepackage{amssymb}
\usepackage{epsfig}
\usepackage{graphicx}
\usepackage{amsmath}
\usepackage{amssymb}
\usepackage{enumitem}
\usepackage[capbesideposition=outside,capbesidesep=quad]{floatrow}
\usepackage{booktabs}
\usepackage{xcolor,colortbl}
\def\eg{\emph{e.g.} }
\def\ie{\emph{i.e.} }

\def\etal{\emph{et al.} }

\newcommand{\boldparagraph}[1]{\vspace{0.3em}\noindent{\bf #1}}
\definecolor{Gray}{gray}{0.9}
\usepackage[breaklinks=true,bookmarks=false]{hyperref}
\iccvfinalcopy
\def\iccvPaperID12 
\def\httilde{\mbox{\tt\raisebox{-.5ex}{\symbol{126}}}}
\ificcvfinal\fi

\begin{document}

\title{Relational Prior Knowledge Graphs for Detection and Instance Segmentation}

\author{Osman Ülger\textsuperscript{1}\quad Yu Wang\textsuperscript{2}\quad Ysbrand Galama\textsuperscript{2}\quad Sezer Karaoglu\textsuperscript{1}\quad Theo Gevers\textsuperscript{1}\quad Martin R. Oswald\textsuperscript{1}\\
\textsuperscript{1}University of Amsterdam\quad \textsuperscript{2} TomTom\\
{\tt\small \{o.ulger,s.karaoglu,th.gevers,m.r.oswald\}@uva.nl}\: {\tt\small \{yu.wang,ysbrand.galama\}@tomtom.com}}

\maketitle
\ificcvfinal\thispagestyle{empty}\fi

\begin{abstract}
Humans have a remarkable ability to perceive and reason about the world around them by understanding the relationships between objects. In this paper, we investigate the effectiveness of using such relationships for object detection and instance segmentation. To this end, we propose a Relational Prior-based Feature Enhancement Model (RP-FEM), a graph transformer that enhances object proposal features using relational priors. The proposed architecture operates on top of scene graphs obtained from initial proposals and aims to concurrently learn relational context modeling for object detection and instance segmentation. 

Experimental evaluations on COCO show that the utilization of scene graphs, augmented with relational priors, offer benefits for object detection and instance segmentation. RP-FEM demonstrates its capacity to suppress improbable class predictions within the image while also preventing the model from generating duplicate predictions, leading to improvements over the baseline model on which it is built.
\end{abstract}

\section{Introduction}
In cognitive psychology, it is well established that humans have a remarkable ability to perceive and reason about the world around them by understanding the relationships between objects~\cite{Bonner2021, highlevescenestatistics, KAISER2019672}. By recognizing how objects relate to each other, humans can build a mental representation of their environment, reason about possible actions and make predictions about outcomes of such actions. This ability is essential for a wide range of tasks, from simple everyday activities like crossing the road to more complex tasks like understanding natural language, planning and decision-making. Similarly, in the field of computer vision, relationships between objects have become an increasingly important topic of research~\cite{reasoningrcnn, hkrm, classificationbyattention}. By leveraging relationships between objects, computer vision systems can enhance their ability to detect and segment objects in images, as well as to reason about their relationships, making it possible to build more sophisticated applications that require a deeper understanding of the visual world. 

\begin{figure}
  \centering
  \includegraphics[width=\columnwidth]{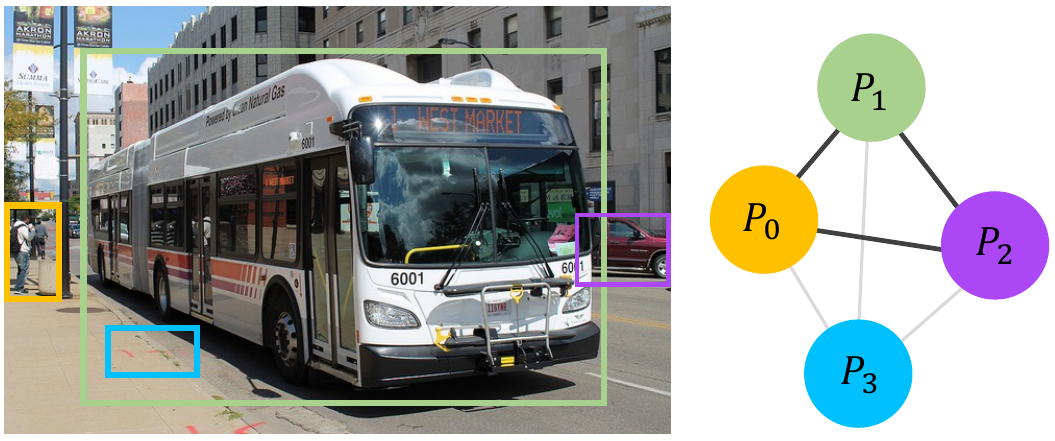}\\
  \caption{\textbf{Problem Setting.} From proposals, a graph is constructed and its edge weights are determined and assigned using prior knowledge. Redundant proposals (\eg in blue) receive low edge weights due to their limited contextual relevance to other objects in the scene. As a result, less contextual feature information from neighboring nodes is aggregated to and from these proposal nodes, which eventually removes them from the final prediction.}
  \label{fig:teaser}
\end{figure}

In this paper, we explore the role of different types of object relationships in instance segmentation, and propose a method for enhancing object proposals by modeling relationships between them. To this end, we introduce a Relational Prior-based Feature Enhancement Model (RP-FEM), a novel framework which combines a multi-headed attention mechanism to select relevant priors and a graph transformer model to aggregate them. Images are represented as scene graphs, where visual feature representations of proposals are modeled as graph nodes and multi-dimensional edges are obtained from prior knowledge about object-object relations. We propose to represent such relations in a Relational Prior Knowledge Graph (RPKG), which is sourced from the scene graph dataset Visual Genome (VG) \cite{visualgenome}. Different from previous works, which enhance proposal features with relational priors for scene graph classification~\cite{classificationbyattention} or object detection~\cite{hkrm, reasoningrcnn}, RP-FEM does not rely on ground truth object regions or an initial classification of proposals. Instead, our method is able to compute relevant relational prior values by attending object neighborhoods in the scene graph with object neighborhoods in the RPKG.

Experimental evaluations on COCO show that using scene graphs, enhanced with relational priors, is beneficial for object detection and instance segmentation. Our method demonstrates its capacity to filter out predicted areas which are improbable when considered in context to other candidate objects in the scene. Furthermore, RP-FEM shows a remarkable capability of reducing the amount of duplicate predictions of the same object instances. All code is publicly available to facilitate additional research on this topic\footnote{https://github.com/ozzyou/RP-FEM}.\\

\noindent
In summary, our contributions are as follows: 
%
\begin{enumerate}
    \item We propose RP-FEM, a novel graph transformer-based model to enhance object proposal features for object detection and instance segmentation using relational priors;
    \item We propose and assess multiple types of relational prior knowledge graphs as relational priors to our model. We demonstrate through qualitative and quantitative evaluations that our model is able to improve in cases where object context is relevant when making predictions.
\end{enumerate}
\section{Related Work}
\boldparagraph{Instance Segmentation.}
Instance segmentation is a challenging task in computer vision that involves identifying and segmenting objects within an image. 
The problem is approached by using convolutional neural networks (CNNs)~\cite{maskrcnn, deepmask}, graph-based methods~\cite{Cao_2019_ICCV, XU2022104739} and transformer-based techniques~\cite{NEURIPS2021_b7087c1f, cheng2021mask2former}. 
Notable works in this field include Mask R-CNN~\cite{maskrcnn}, GCNet~\cite{Cao_2019_ICCV} or Mask2Former~\cite{cheng2021mask2former}. 
Building on top of such architectures, various works are proposed to utilize prior knowledge in order to enable targeted object search~\cite{semisupinstance}, weak supervision~\cite{weaksupbbtightness, weaklymultiprior} or mask refinement~\cite{shapepriorgraph, connectivityprior} with priors originating from bounding box tightness~\cite{weaksupbbtightness, weaklymultiprior}, object shapes~\cite{semisupinstance, shapepriorgraph}, object contours~\cite{weaklymultiprior} or object connectivity~\cite{connectivityprior}.
Despite these efforts to provide instance segmentation algorithms with more object-specific information with priors, leveraging prior knowledge about object-to-object information still remains an open problem. Although the majority of instance segmentation models benefit from spatial context information in feature space, explicit relationships between objects are ignored. 

In contrast to existing methods, we propose to utilize such contextual information by modeling common knowledge about relations between objects in an image, which can constitute to multiple relation types. We extend the Mask R-CNN architecture and showcase its advantages through the integration of relational priors, which can be readily derived from existing dataset statistics.

\boldparagraph{Feature Enhancement with Relational Priors.}
Relational reasoning for feature enhancement is studied in different areas. 
Kang et al.~\cite{grn} propose a graph relation network which embeds more discriminative metric spaces for image classification. 
Relation distillation networks \cite{relationdistillation} improve video object detection by modelling appearance and geometric relations via multi-stage reasoning. 
While these works fall under a collection of works which utilize relationships through combination of instance features \cite{relationnetworkobjectdetection}, a large cohort of other works focus on reasoning about scenes by detecting and classifying relationships from visual input \cite{detectingvisualrel, graphcontrasticescenegen}, driven by scene graph datasets Visual Relation Detection (VRD) \cite{VRD} and Visual Genome (VG)~\cite{visualgenome}. 
The creation of VG enabled the development of models which use object-to-object relations, often co-occurrence statistics, as external high-level knowledge, which can in turn be used when reasoning about scenes.
In previous works, relational prior knowledge about object-object relations is exclusively applied to scene graph classification \cite{classificationbyattention}, object classification, \cite{classificationbyattention, hkrm, reasoningrcnn} or object detection \cite{hkrm, reasoningrcnn}. 

Sharifzadeh~\etal\cite{classificationbyattention} show that prior knowledge provides significant improvements in the scene graph and object classification tasks. However, their method relies on available ground truth bounding boxes during inference and therefore a detection or segmentation model cannot be trained end-to-end. For object detection, Jiang~\etal\cite{hkrm} consider edges in the prior knowledge graph as a supervision signal for edges in the scene graph, predicted from the pairwise L1 difference between the features of each region pair. In our work, the relational prior knowledge is obtained via attention over pairs of readily available object proposal features without loss of feature information. 
Xu~\etal~\cite{reasoningrcnn} propose Reasoning-RCNN in which a category-to-category undirected graph is constructed from classifier weights. Region proposals are then mapped to each class node, essentially forming an initial classification. This graph is then evolved to obtain contextualized class embeddings, which are then concatenated to region proposals. 

Our work does not rely on initial classifications and operates fully in the region proposal space, thereby limiting the chance of incorporating wrong prior knowledge as a consequence of misclassification. We furthermore apply our method to instance segmentation, while existing related works exclusively focus on scene graph classification, object classification and object detection. Obtained relational priors represent the importance of each object-object relation in a scene before proposal features are propagated across a fully connected and directed graph.

\section{Method}
Our aim is to enhance proposal features of an underlying base detection and instance segmentation model - the Mask R-CNN framework in this paper - with relational prior knowledge using a Relational Prior Knowledge Graph (RPKG) and \textit{context updates}. In this section, we first provide details on how a RPKG is constructed and which relation types are considered in our proposed architecture. Then, we introduce our RP-FEM model to predict edge weights, representing object-object relations, in a fully-connected scene graph using the RPKG. Next, we detail how enhanced proposal features are obtained from the scene graph through context updates. Lastly, we provide details on how the enhanced proposal features are adapted within the Mask R-CNN framework.

\begin{figure*}[t]
  \centering
  \includegraphics[width=\linewidth]{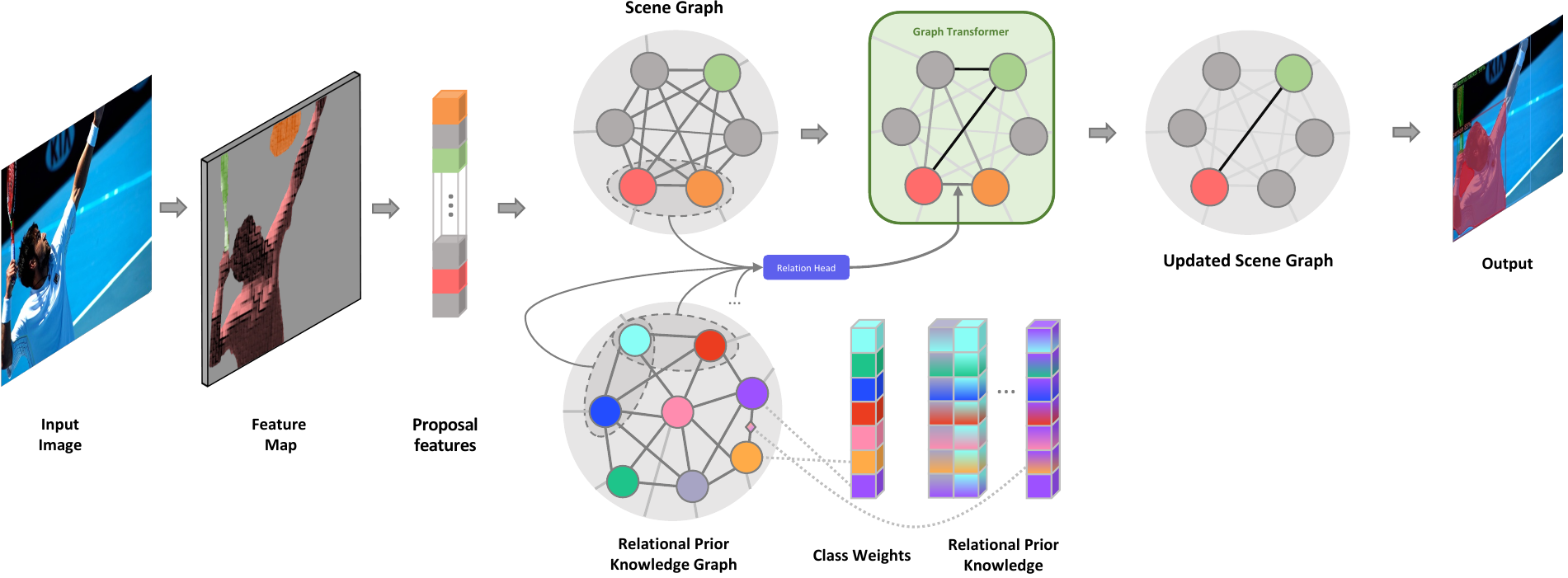}\\
  \caption{\textbf{Method overview.} Our Relation Head updates each edge in the Scene Graph with relational prior knowledge by attending node neighborhoods in the Scene Graph (representing  proposals) with node neighborhoods in the Relational Prior Knowledge Graph (representing class embeddings). Original proposal features and predicted edges are fed to a Graph Transformer to obtain an updated Scene Graph. From the updated Scene Graph, bounding boxes and masks are predicted.}
  \label{fig:approach_overview}
\end{figure*}

\subsection{Relational Prior Knowledge Graph} \label{relpriorgraph}
Before training, we build three different Relational Prior Knowledge Graphs (RPKG) from the Visual Genome (VG) \cite{visualgenome} dataset once.
As the nodes of the RPKG, we use the feature representation $d \in \mathbf{D}^{C \times F}$ of each class $c \in \textbf{C}$ in the penultimate layer of a Faster R-CNN model \cite{fasterrcnn} pre-trained on $C$ classes with $F$ feature dimensions. We use Faster R-CNN due to its architectural similarity with Mask R-CNN regarding object detection.
Using the scene graph annotations in VG, we collect object-object relationships which represent the edges of the RPKG. The different relationship types consist of:
\begin{enumerate}[itemsep=0.1pt,topsep=3pt,leftmargin=*,label=\textbf{(\arabic*)}]
    \item \textbf{Co-occurrence.} Measures how often on average two object classes appear together across the dataset. For each class, the amount of co-occurence with other classes is collected and divided by the individual appearances in a scene.
    \item \textbf{Relative Orientation.} When two objects $\{A, B\}$ appear in the same scene, the relative orientation measures how often object $A$ is \textit{at the center of, left of, right of, above or below} object $B$. 
    Multiple options, such as \textit{left of} and \textit{above}, can occur at the same time, \ie "A is above and left of B". 
    For each object pair, the 5-dimensional outcome is averaged over all samples in which the objects co-occur.
    \item \textbf{Relative Distance.} Measures the mean distance and mean standard deviation between the locations of two objects which co-occur in an image, relative to the size of its ground truth bounding box and size of the image.
\end{enumerate}
This results in the relational prior knowledge graph  $\textbf{R} = \langle \mathbf{D},\mathbf{K}\rangle$, with $\mathbf{K} \in \mathbb{R}^{C \times C \times R}$ where $R$ depends on which relations are used. We consider relationships between object categories in VG which overlap with the COCO classes. However, some object category names in VG do not have a one-to-one correspondence with those in COCO, \eg ``hair blower" in VG versus ``hair dryer" in COCO. In order to mitigate this, we manually link semantically identical object categories. The RPKGs, code to construct them and the dataset-to-dataset mappings of object categories will be made publicly available upon publication.

\subsection{From Prior Knowledge to Useful Knowledge} \label{relprior}
Our method is designed to enhance proposal features with prior knowledge information. Central to our proposed architecture is the process of retrieving relevant relational information from $\textbf{R}$ based on the appearance of potential objects in the proposals and to represent the relational information in the proposal feature space. Naturally, proposal features do not consider the relations among them. In order to do so, we first construct a scene graph $\mathbf{S}= \langle \mathbf{P},\mathbf{E}\rangle$ given a set of proposal features $\{p_i, ..., p_\mathcal{N}\} = \mathbf{P} \in \mathbb{R^{\mathcal{N} \times \mathcal{F}_{\text{p}}}}$ representing the nodes and a set of edges $\{e_{ii}, e_{ij}, ..., e_{\mathcal{N}\mathcal{N}}\} = \mathbf{E} \in \mathbb{R^{\mathcal{N}\times \mathcal{N} \times \mathcal{F}_{\text{e}}}}$. In the next step, we predict $\mathbf{E}$ using $\mathbf{P}$ and $\textbf{R}$, thereby retrieving relevant relational information.

To retrieve this relational information, the aim is to compute the similarity between pairs of node features - or neighborhood features in $\mathbf{S}$ - and pairs of node features in $\textbf{R}$ using an attention mechanism $\mathrm{att}(\cdot)$. Weighted by the extent of similarity between two node neighborhoods, the edge value of each neighborhood in $\textbf{R}$ is aggregated. More formally, the attention coefficient $\alpha_{(ij), (uv)}$ is computed between each pair of nodes $[p_i, p_j] \in \mathbf{P}$, representing the queries, and all pairs of nodes $[d_u, d_v] \in \textbf{R}$, representing the keys. Features of node neighborhoods are stacked and linearly transformed with shared weight matrices to create local, latent neighborhood representations $\hat{p}_{ij} \in \mathbb{R^{\mathcal{F}_{\text{p}} + \mathcal{F}_{\text{p}}}}$ and $\hat{d}_{uv} \in \mathbb{R^{\mathcal{F}_{\text{r}} + \mathcal{F}_{\text{r}}}}$ for $\mathbf{S}$ and $\textbf{R}$ respectively. In order to compute the final edge values $\mathbf{E}$ in $\mathbf{S}$, linearly transformed edge values in $\textbf{R}$, representing the values, are multiplied by the corresponding attention weights:
\begin{align}
    \alpha_{(ij),(uv)}
    &= \frac{\exp\big(\mathrm{att}(\mathbf{W}_q[p_i, p_j], \mathbf{W}_k[d_u, d_v])\big)}{\sum\limits_{u=0}^C\sum\limits_{v=0}^C\exp\big(\mathrm{att}(\mathbf{W}_q[p_i, p_j], \mathbf{W}_k[d_u, d_v])\big)} \nonumber\\
    &= \frac{\exp\big(\mathrm{att}(\hat{p}_{ij}, \hat{d}_{uv})\big)}{\sum\limits_{u=0}^C\sum\limits_{v=0}^C\exp\big(\mathrm{att}(\hat{p}_{ij}, \hat{d}_{uv})\big)} %
\end{align}
\begin{align}
  e_{(ij),(kl)}   &= \alpha_{(ij),(kl)})\mathbf{W}_v\text{R}_{kl} \\
  \mathbf{E}_{ij} &= \mathbf{W}_{\text{E}}\sum^C_{k=0}\sum^C_{l=0}e_{(ij),(kl)}
\end{align}
where $\mathbf{W}_q, \mathbf{W}_k, \mathbf{W}_v, \mathbf{W}_{\mathbf{E}}$ represent shared weight matrices to obtain latent representations of the queries, keys, values and predicted edge values respectively. The resulting edge matrix $\mathbf{E}$ now weighs the importance of a proposed object to other proposed objects in the scene graph based on relational prior knowledge obtained from occurrences (or lack of occurrences) of such object combinations in the prior knowledge graph. By using proposals to obtain the prior-knowledge based edge matrix $\mathbf{E}$ with attention, our method does not rely on the classification of proposals and therefore avoids possible challenges posed when proposals are incorrectly classified.

\subsection{Context Update}
After predicting all edge values in $\mathbf{E}$ of the scene graph $\mathbf{S}$, we execute an operation referred to as the \textit{context update}, employing a multi-layered Graph Transformer \cite{relindbiases, textgen, classificationbyattention}. The context update ensures that node features in $\mathbf{S}$ are aggregated across the graph to provide each node with more context about the entire scene, as well as the usual, prior-based relations it has with other nodes in such a context. In the process, each node gets informed about its neighboring nodes through messages $\mathbf{m}$, weighted by the edge matrix $\mathbf{E}$ or $\mathbf{A}$, to result in context-aware nodes $\{\mathbf{z}_i, ..., \mathbf{z}_\mathcal{N}\} = \mathbf{Z} \in \mathbb{R}^{\mathcal{N} \times \mathcal{F}_{\text{z}}}$:
\begin{equation} \label{eq:f}
    \mathbf{f}_{ij}^{(l)} = \mathcal{E}(\delta_{ij}))
    \qquad \text{with} \quad
    \delta_{ij} = \begin{cases}
        \mathbf{E}_{ij} & \text{if } l=0 \\
        \mathbf{A}_{ij}^{(l-1)} & \text{if } l>0
    \end{cases}
\end{equation}
\begin{equation} \label{eq:f}
    \mathbf{n}_{i}^{(l)} = \gamma_{ij}
    \qquad \text{with} \quad
    \gamma_{ij} = \begin{cases}
        p_i & \text{if } l=0 \\
        \mathbf{z}_i^{(l-1)} & \text{if } l>0
    \end{cases}
\end{equation}
\begin{align}
    \alpha_{ij}^{(l)} &= \sigma\big(\mathrm{LReLU}([\mathbf{f}_{ij}^{(l)} \oplus \mathbf{n}_i^{(l)}])\big)  \\
    \mathbf{m}_i^{(l)} &= \sum_{j \in \mathcal{I}} \alpha_{ij}^{(l)} \mathbf{f}_{ij}^{(l)}  \\
    \hat{\mathbf{z}}_i^{(l)} &=\mathbf{LN}\left(\mathbf{n}_i^{(l)}+\mathbf{m}_i^{(l),\text{head}}+\mathbf{m}_i^{(l),\text{tail}}\right)  \\
    \mathbf{z}_i^{(l)} &=\mathbf{LN}\left(\hat{\mathbf{z}}_i^{(l)}+f\big(\hat{\mathbf{z}}_i^{(l)}\big)\right)  \label{eq:z}
\end{align}
where $\mathcal{F}_{\text{z}}$ represents the dimension of the output features, $l$ represents the $l$-th Graph Transformer layer, $\mathcal{E}$ is the transformer function applied to the edge features, $\sigma(\cdot)$ the $\mathrm{Softmax}$ function, $\mathrm{LReLU}$ the Leaky ReLU activation function \cite{leakyrelu}, $\oplus$ is concatenation, $\mathbf{A}$ the updated adjacency matrix (explained in Eqs.~\eqref{eq:h}-\eqref{eq:A}), $\mathbf{LN}$ the LayerNorm operation \cite{layernorm} and $f(\cdot)$ consists of two linear layers with a Leaky ReLU after each layer. The first Graph Transformer layer considers the original edge matrix $\mathbf{E}$, whereas in following layers the edge matrix is updated to give $\mathbf{A}$:
\begin{align}
    \mathbf{h}_i^{(l),\text{head}} = \mathcal{H}(\mathbf{n}_i^{(l)}) \qquad\qquad
    \mathbf{h}_i^{(l),\text{tail}} = \mathcal{T}(\mathbf{n}_i^{(l)})  \label{eq:h}
\end{align}  
\begin{align}
    \begin{split}
    \alpha_i^{(l),\text{head}} &= \mathrm{LReLU}\big(\mathcal{A}([\delta_i^{\text{head}} \oplus \mathbf{h}_i^{(l),\text{head}}])\big)  \\
    \alpha_i^{(l),\text{tail}} &= \mathrm{LReLU}\big(\mathcal{A}([\delta_i^{\text{tail}} \oplus \mathbf{h}_i^{(l),\text{tail}}])\big)  \\
    %
    \text{with} \quad
    \delta_i &= \begin{cases}
        \mathbf{E}_i & \text{if } l=0 \\
        \mathbf{A}_i^{(l-1)} & \text{if } l>0
    \end{cases}
    \end{split}
\end{align}
\begin{align}
    \alpha_i^{(l),\text{head+tail}} &= \sigma([\alpha_i^{(l),\text{head}} \oplus \alpha_i^{(l),\text{tail}}])\\
    \mathbf{A}_i^{(l)} &= \alpha_i^{(l),\text{head+tail}} \odot [\mathbf{h}_i^{(l),\text{head}} \oplus \mathbf{h}_i^{(l),\text{tail}}] \label{eq:A}
\end{align}
where additionally $\mathcal{H}$ and $\mathcal{T}$ are the transformer functions applied to proposal features with head or tail indices respectively, $\mathcal{A}$ is the transformer function applied to the concatenated node and edge features and $\odot$ represents element-wise multiplication. A Leaky ReLU activation function is applied to the concatenated features to enable non-linearity. Next, a $\mathrm{Softmax}$ layer is applied on the stacked attention coefficients for the head and tail indices. Finally, the adjacency matrix $\mathbf{A}_i^{(l)}$ is obtained by multiplying the attention coefficients with the transformed proposal features from the head and tail indices, which can then be utilized to properly weigh node features during the feature aggregation step described in Eqs.~\eqref{eq:f}-\eqref{eq:z}.

\subsection{Mask Prediction}
After $L$ iterations of context updates, the edge matrix is discarded and the final, $L$-th node features $\mathbf{Z}^{(L)}$ are concatenated with the original proposal features $\mathbf{P}^{box}$ to yield output features $\mathbf{O}^{box}$ used for bounding box prediction. The output features $\mathbf{O}^{mask}$ for mask prediction are obtained after concatenation with proposal features $\mathbf{P}^{mask}$ which contain foreground objects found by the box head $\mathcal{B}$. More specifically:
\begin{align}    
    \mathbf{O}^{box}  &= [\mathbf{P}^{box} \oplus \mathbf{Z}] \\
    \mathbf{O}^{mask} &= [\mathbf{P}^{mask}_{\mathcal{B}(\mathbf{O}_{bb})} \oplus \mathbf{Z}] = [\mathbf{P}^{mask}_{fg} \oplus \mathbf{Z}]
\end{align}

We ensure that the Mask R-CNN framework is adapted to the shape size increase as a result of the concatenations by adjusting the accepted input size of the box and mask head. Since the prior knowledge graph is fixed, it does not have a gradient and is thus not trained. This allows for training on the image data (COCO) alone. The final output is supervised end-to-end with Mask R-CNN's original loss function, consisting of a term for each of the predictions:
\begin{equation}
    \mathcal{L} = \mathcal{L}_{cls} + \mathcal{L}_{box} + \mathcal{L}_{mask}.
\end{equation}
\section{Experiments}
\subsection{Experimental Setup}
\boldparagraph{Datasets and Classes.}
We obtain the relational priors from ground-truth annotations of the scene graph edges in Visual Genome (VG) \cite{visualgenome}, which contains over 3000 object categories. However, since VG is not a segmentation dataset, we instead train and evaluate our model on the Common Objects in Context (COCO) \cite{cocodataset} dataset which contains 80 classes. Both VG and COCO are licensed under a Creative Commons Attribution 4.0 License. Naturally, VG contains many redundant classes when training and evaluating on COCO, posing the need to select only overlapping classes. Therefore, we automatically assign VG labels to COCO labels which are semantically identical based on WordNet synsets~\cite{wordnet} and extract the relational prior knowledge for a class in COCO from all VG classes that have been assigned to it. This results in relational prior knowledge for all possible classes in COCO.

\boldparagraph{Training.}
To compare in a fair manner, we set the training parameters similar to \cite{maskrcnn}. Our models are trained on 8 Tesla V100 GPUs with 32GB memory for 90k iterations with a batch size of 16. The Mask R-CNN model we extend using RP-FEM uses a ResNet-50-FPN backbone and ROIAlign. The feature dimensions $\mathcal{F}_{\text{p}}$, $\mathcal{F}_{\text{r}}$ and $\mathcal{F}_{\text{z}}$ for the node, edge and updated node features are set to 1024 latent dimensions respectively. To sustain a memory-efficient scene graph, we experiment with 128 or 448 proposals originating from the Region Proposal Network (RPN), different from the default 512 in Mask R-CNN. During all experiments, we train the box- and mask head concurrently.
\subsection{Ablation Studies}
In this section, we present three ablation studies to investigate the preferred settings for the amount of relation heads, context updates and effect of each relation type. In these initial experiments, 128 proposals are used and results are reported for instance segmentation.

\boldparagraph{Relation Heads.} Having multiple attention heads, \ie attention mechanisms, has been shown to stabilize the learning process in attention-based approaches \cite{attentionallyouneed, GAT}. Therefore, we first investigate the effect of the number of attention heads on the performance of RP-FEM. In this initial experiment, we set the amount of relation heads to 1, 2 or 4. Results are presented in Table~\ref{tab:ablationssegpred} and demonstrate that having multiple relation heads positively impacts the performance when compared to a single head. Given our setting, one potential explanation for the effectiveness of employing multiple heads is that it allows for learning diverse dynamics within a single layer. Interestingly, we observe that 4 relation heads allow the model to more accurately segment instances of small and medium size, while overall, 2 relation heads are preferred. This could indicate that small and medium-sized instances benefit from more dynamics captured by more attention heads. Throughout following experiments, we employ 2 relation heads.

\begin{table}[tb]
\centering
\scriptsize
\setlength{\tabcolsep}{3.5pt}
\renewcommand{\arraystretch}{1.2}
\newcommand{\csp}{\hskip 2em}
\begin{tabular}{l ccccccc}
& & $AP\!\uparrow$ & $AP_{50}\!\uparrow$ & $AP_{75}\!\uparrow$ & $AP_{s}\!\uparrow$ & $AP_{m}\!\uparrow$ & $AP_{l}\!\uparrow$\\
\rowcolor{gray!20}
\rowcolor{gray!20}
\multicolumn{8}{l}{\textbf{Relation Attention Heads}} \vspace{2px}\\
1 & & 33.11 & 52.33 & 35.65 & 15.09 & 34.97 & 50.10 \\
2 & & \textbf{33.75} & \textbf{53.17} & \textbf{36.43} & 14.85 & 35.56 & \textbf{51.52} \\
4 & & 33.62 & 53.06 & 36.14 & \textbf{15.90} & \textbf{35.78} & 50.45 \vspace{1px}\\
\rowcolor{gray!20}
\multicolumn{8}{l}{\textbf{Context Updates}} \vspace{2px}\\
1 & & \textbf{33.75} & \textbf{53.17} & \textbf{36.43} & 14.85 & \textbf{35.56} & \textbf{51.52} \\
2 & & 33.24 & 52.87 & 35.75 & \textbf{15.70} & 35.24 & 49.37 \\
3 & & 32.73 & 52.35 & 35.02 & 15.09 & 34.57 & 49.11 \vspace{1px}\\
\rowcolor{gray!20}
\multicolumn{8}{l}{\textbf{Relationship Types}} \vspace{2px}\\
Co-occurence & & 33.88 & \textbf{53.70} & 36.40 & 15.54 & 35.75 & 51.14 \\
Relative Orientation & & 33.71 & 53.33 & 36.04 & 15.43 & 35.59 & \textbf{51.58} \\
Relative Distance & & \textbf{33.90} & 53.56 & \textbf{36.65} & \textbf{15.56} & \textbf{36.16} & 51.29 \\
All & & 33.75 & 53.17 & 36.43 & 14.85 & 35.56 & 51.52 \\
\end{tabular}
\caption{\textbf{Overview of all ablation studies on COCO Instance Segmentation} \cite{cocodataset}. Our findings provide the following insights: i) two relation attention heads aid learning best through added complexity; ii) one graph transformer layer is all you need; and iii) overall, Relative Distance proves as the strongest relational prior for instance segmentation.}
\label{tab:ablationssegpred}
\end{table}

\begin{table*}[tb]
\centering
\small
\setlength{\tabcolsep}{8pt}
\renewcommand{\arraystretch}{1.2}
\newcommand{\csp}{\hskip 6em}
\begin{tabular}{l@{\csp}c@{\csp}ccc@{\csp}ccc}
 & \multicolumn{1}{c}{ } & \multicolumn{3}{c@{\csp}}{\textbf{Object Detection}} & \multicolumn{3}{c}{\textbf{Instance Segmentation}} \vspace{3px} \\
& Proposals & $AP$ $\uparrow$ & $AP_{50}$ $\uparrow$ & $AP_{75}$ $\uparrow$ & $AP$ $\uparrow$ & $AP_{50}$ $\uparrow$ & $AP_{75}$ $\uparrow$\\
\rowcolor{Gray}
\multicolumn{8}{l}{\textbf{Previous Work}} \vspace{2px}\\
Mask R-CNN~\cite{maskrcnn} & 512 & 38.5 & 59.1 & 42.0 & 35.0 & 56.0 & 37.5 \\
GCNet~\cite{Cao_2019_ICCV} & 2000 & 37.2 & 59.0 & 40.1 & 33.8 & 55.4 & 35.9 \\
MS R-CNN~\cite{maskscoringrcnn} & 2000 & \textbf{38.6} & \textbf{59.2} & \textbf{42.5} & \textbf{36.0} & \textbf{55.8} & \textbf{38.8} \vspace{1px}\\
\rowcolor{Gray}
\multicolumn{8}{l}{\textbf{Baseline Improvement}}\vspace{2px} \\
Mask R-CNN~\cite{maskrcnn} & 128 & 36.7 & 56.3 & 40.3 & 33.5 & 53.7 & 35.7 \\
RP-FEM (Ours) & 128 & 36.2 & 55.8 & 39.8 & 33.9 & 53.6 & 36.7 \\
Mask R-CNN~\cite{maskrcnn} & 448 & 38.4 & 59.2 & 42.0 & 35.1 & 56.1 & 37.6 \\
RP-FEM (Ours) & 448 & \textbf{38.7} & \textbf{59.4} & \textbf{42.2} & \textbf{35.3} & \textbf{56.4} & \textbf{37.8} \\
\end{tabular}
\caption{\textbf{Quantitative Results of Object Detection and Instance Segmentation on COCO val2017} \cite{cocodataset}. RP-FEM succesfully enhances the performance of the Mask R-CNN baseline with relational priors. With considerably less proposals, we also outperform GCNet \cite{Cao_2019_ICCV}, which similarly tries to model global context for feature enhancement.}
\label{tab:sotaAG}
\end{table*}

\boldparagraph{Context Updates.} The RP-FEM architecture allows for iterative aggregation of node features, referred to as context updates, using relational priors as edges. This architecture enables the model to enhance scene graph nodes with increased contextual information at each progressive layer. The node and edge features from previous graph transformer layers are propagated to subsequent layers. To determine the optimal number of context updates, this ablation study examines the results for the COCO dataset, as depicted in Table~\ref{tab:ablationssegpred}. Interestingly, the best performance is achieved with a single context update across most metrics, except for $AP_{s}$, where two updates are preferable and result in a significant performance gap. From these findings, we can conclude that it is more important to richly model a single graph transformer layer with multiple relation heads rather than modeling higher-order neighborhood context. For this reason, 1 graph transformer layer is modeled in following experiments.

\boldparagraph{Relationship Types.} In the final ablation study, we conduct an evaluation of the performance of each individual relationship type. The results, presented in Table~\ref{tab:ablationssegpred}, reveal that specific relationship types offer advantages for different metrics. Notably, the Relative Distance relationship type exhibits superior performance across all metrics, with an average AP of 33.904. This outcome is surprising considering that the literature often uses co-occurrence metrics~\cite{hkrm, reasoningrcnn}. Co-occurrence proves more effective when computing AP with an Intersection over Union (IoU) threshold of 50\%. This suggests that when considering a larger number of proposal predictions, co-occurrence enhances more proposals successfully compared to other relationship types. Moreover, Relative Orientation appears to play a crucial role in the case of large objects. Interestingly, an ensemble of edge features incorporating all relationship types does not yield the best results for any of the metrics. This observation indicates the potential for exploring additional individual relationship types in future studies.

\begin{figure}[b]
  \centering
  \includegraphics[width=\columnwidth]{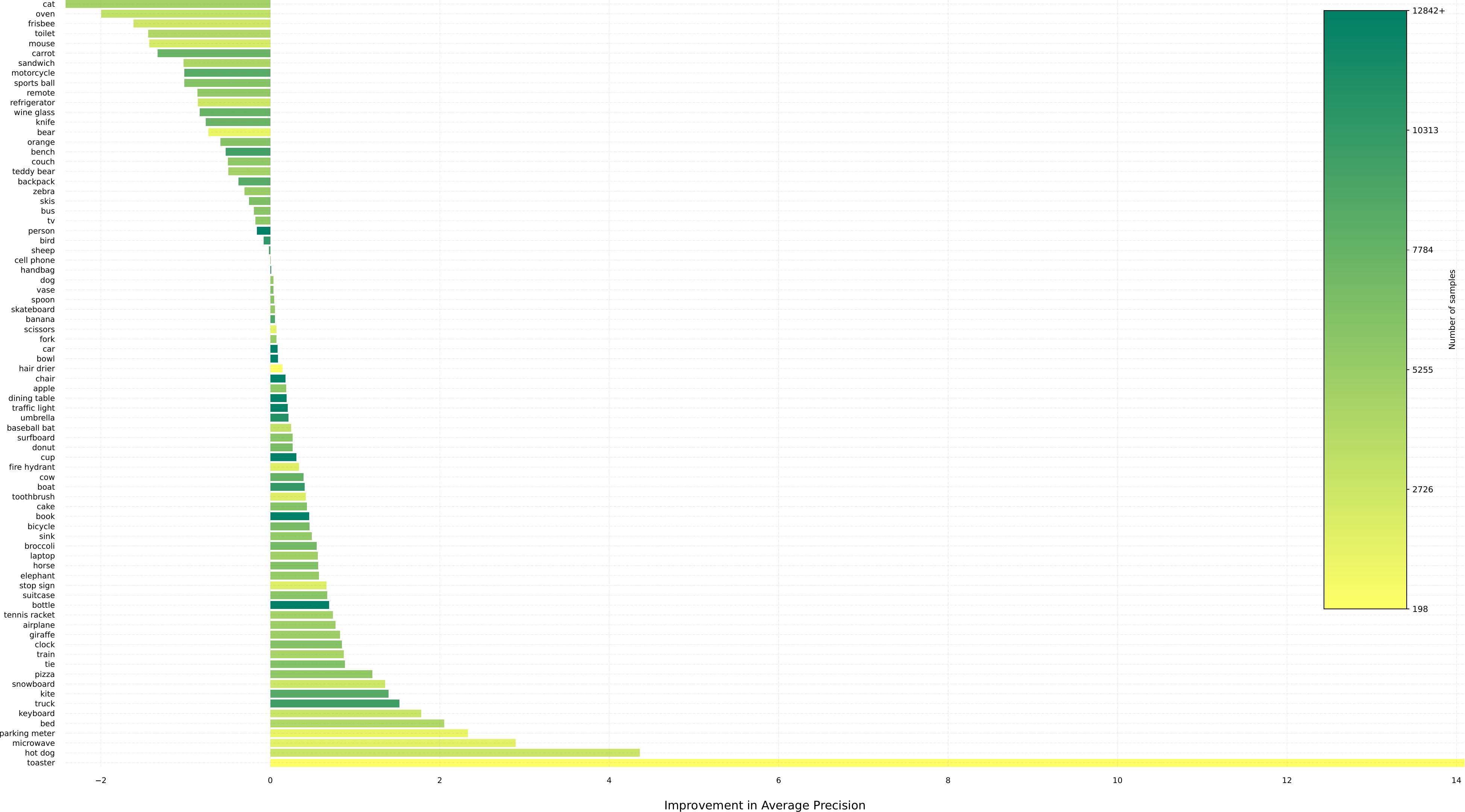}\\
  \caption{\textbf{AP improvement per class.} RP-FEM improves on two thirds of the classes in COCO over Mask R-CNN~\cite{maskrcnn}. Classes with a low number of samples in the dataset particularly benefit from relational prior knowledge.}
  \label{fig:barplot}
\end{figure}

\begin{figure*}[t]
  \centering
  \includegraphics[width=\linewidth]{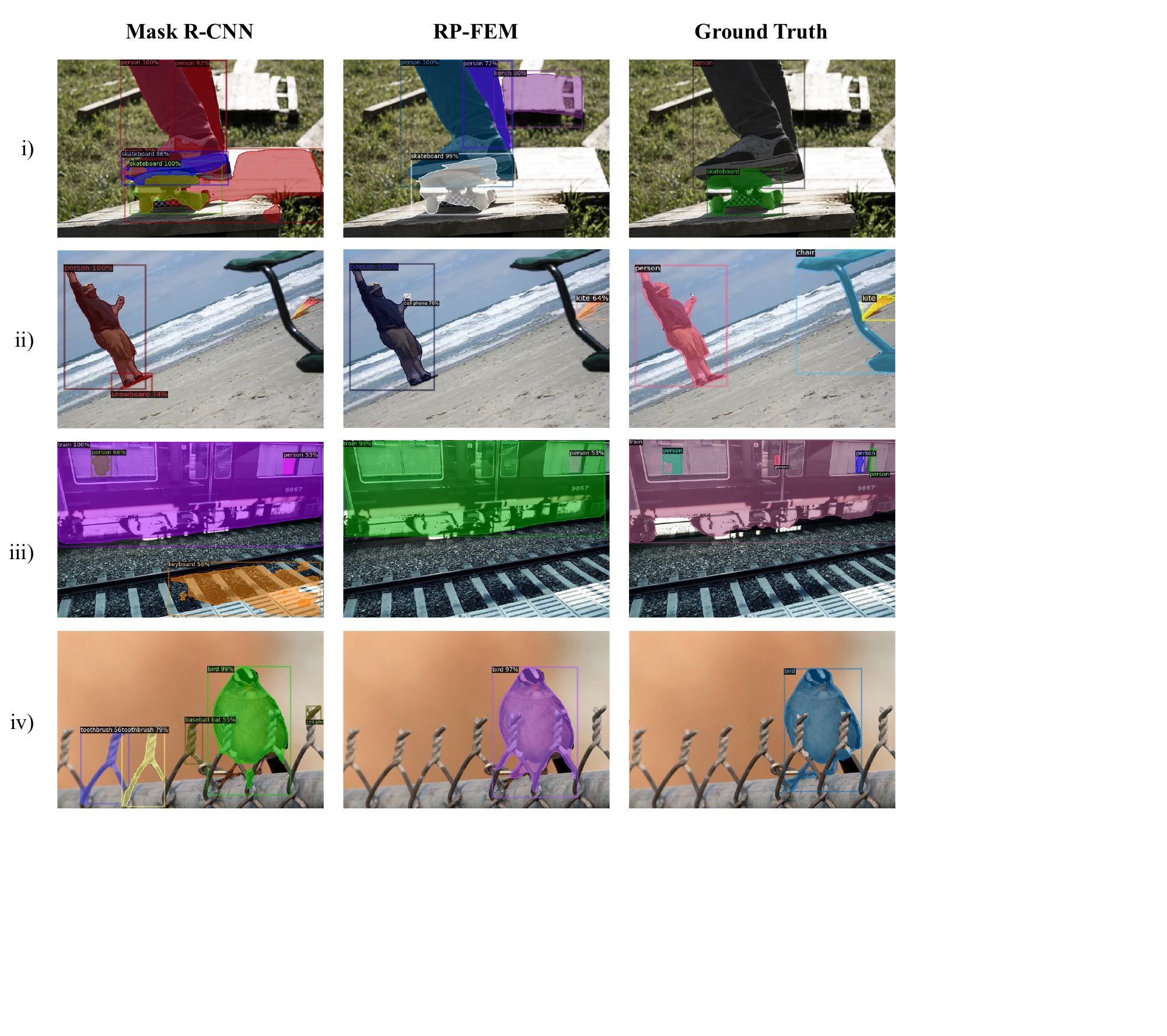}\\
  \caption{\textbf{Qualitative results on COCO \cite{cocodataset}} with the following observations: i) our model is able to more accurately predict the correct number of instances, while Mask R-CNN creates duplicate predictions; ii) thanks to contextual information, our model is better able to suppress objects (snowboard) which likely occur with one object (the person), but unlikely in context of multiple objects (the kite). One adverse side effect of relational prior knowledge is the hallucination of objects, such as the cell phone, likely due to many co-occurrences of person and cell phone; iii) RP-FEM filters regions that have visual similarities with other objects (railway as keyboard) if the context does not make sense; iv) multiple instances can be suppressed at once. }
  \label{fig:qualitative}
\end{figure*}

\subsection{Quantitative Analysis}
In Table~\ref{tab:sotaAG}, we provide quantitative results of RP-FEM in comparison to previous works which utilize a similar ResNet-50 backbone and two-stage framework for object detection and instance segmentation. We note that previous works such as GCNet~\cite{NEURIPS2021_b7087c1f} and MS R-CNN~\cite{maskscoringrcnn} require a large number of object proposals in order to achieve the reported performance, and comparing against them would require us to reduce the amount by a factor of over 15. Hence, we mainly compare against Mask R-CNN with the number of proposals set to either 128 or 448. When we compare RP-FEM in the object detection task with 128 proposals, we observe that it achieves competitive performance but struggles to outperform Mask R-CNN. In the instance segmentation task, however, RP-FEM performs better overall with an average precision score of 33.9 in comparison to 33.5 achieved by Mask R-CNN. This is likely attributed to lower recall when a small number of proposals is used, consequently allowing less context to be propagated across the scene graph. When the number of proposals is increased to 448, we observe a consistent performance increase over Mask R-CNN - even the original version with 512 proposals - across all metrics for both object detection and instance segmentation. GCNet similarly tries to model global context for feature enhancement. This is achieved by applying self-attention on query positions within the image. Our prior-based attention mechanism achieves a superior performance of 35.2 AP over GCNet's 33.8 AP with less than a quarter of the number of proposals (448 versus 2000), while performing on par with MS R-CNN. 

In Figure~\ref{fig:barplot}, we also report the AP improvement per class of RP-FEM over Mask R-CNN. Our model improves the AP of two thirds of the classes in COCO. Classes which have a low number of samples, such as ``toaster", particularly benefit from the incorporation of relational prior knowledge. This indicates that RP-FEM can serve as a promising approach for applications with a long-tailed class distribution. Our analysis indicates the effectiveness of proposal feature enhancement with relational priors and modeling global context.

\subsection{Qualitative Analysis}
A qualitative analysis of our model's performance on the COCO dataset yields insightful observations, highlighted in Fig.~\ref{fig:qualitative}, which can be summarized as follows:

\boldparagraph{Proposal suppression using context.} 
Our model leverages contextual information to effectively suppress objects that are likely to appear in the context of a single object but are improbable when multiple objects are present. This contextual awareness empowers our model to make more informed and contextually consistent predictions. Moreover, our model exhibits a remarkable ability to identify and filter out incorrectly predicted regions that bear visual similarities to certain objects. For example, in Figure~\ref{fig:qualitative}, the third row depicts a region containing a railway that Mask R-CNN mistakenly identifies as a keyboard due to their visual resemblance. In contrast, RP-FEM successfully discards this region, ensuring that predictions align with the logical context of objects in the scene. Additionally, our model showcases the capability to suppress multiple instances simultaneously, enhancing the instance segmentation process. This efficiency leads to more accurate outputs.

\boldparagraph{Accurate instance count prediction.} Our model demonstrates improved accuracy in predicting the correct number of instances compared to Mask R-CNN, likely guided by the co-occurrence relational prior knowledge. Unlike Mask R-CNN, our model avoids generating duplicate predictions better, resulting in a more precise instance segmentation output. For example, in Figure~\ref{fig:qualitative}, Mask R-CNN predicts multiple instances of the skateboard, while RP-FEM correctly identifies one skateboard only.

Through this qualitative analysis, we highlight the strengths of our model in accurately predicting instance counts, leveraging contextual information, filtering regions with conflicting visual similarities, and efficiently suppressing multiple instances. These findings showcase the advancements and superior performance achieved by our approach on the COCO dataset.

\subsection{Limitations}
Our model has the following limitations.
When the number of proposals and/or classes in the relational prior knowledge graph grows, predicting each edge in the scene graph becomes costly in terms of memory consumption. This can be addressed by computing the edges sparsely or iteratively, but both workarounds have their price in accuracy or computation time. 
Furthermore, the incorporation of relational prior knowledge can cause the detection or segmentation model to hallucinate objects.
In Figure~\ref{fig:qualitative}, for example, RP-FEM hallucinates a cell phone in the hand of the man, likely due to many co-occurrences of both objects.
\section{Conclusion}
The understanding of relationships between objects play an important role in human perception and reasoning. In this work, we explored whether utilizing relationships can play a beneficial role in the tasks of object detection and instance segmentation. To this end, we proposed a Relational Prior-based Feature Enhancement Model which employs the unique capability of suppressing multiple region proposals when they present themselves in a context that is unlikely to be consistent. Furthermore, our model provides a more accurate notion of instance counts, reducing the amount of duplicate object detections and instance segmentations around the same object. Our quantitative results further confirm that our model can outperform its base model, as well as comparative models which model context, all the while using less object proposals. We find that, in particular, classes with a low number of samples benefit strongly from the incorporation of relational prior knowledge. We encourage future work to explore linguistic relationships and to experiment with stronger backbones.

{\small
\bibliographystyle{ieee_fullname}
\bibliography{references}
}

\end{document}